# Machine Learning Based Missing Data Imputation in Categorical Datasets

MUHAMMADISHAQ1,∗, SANA ZAHIR1,∗, LAILA IFTIKHAR1, MOHAMMAD FARHAD BULBUL SEUNGMIN RHO 3,ANDMIYOUNGLEE 4,(Member, IEEE)

1Institute of Computer Sciences and Information Technology, The University of Agriculture at Peshawar, Peshawar, Khyber Pakhtunkhwa 25000, Pakistan 2Department of Mathematics, Jashore University of Science and Technology, Jashore 7408, Bangladesh 3Department of Industrial Security, Chung-Ang University, Seoul 06974, South Korea 4Department of Research, Chung-Ang University, Seoul 06974, South Korea Corresponding author: Mi Young Lee (miylee@cau.ac.kr) 2, This work was supported by the Basic Science Research Program through the National Research Foundation of Korea (NRF) funded by the Ministry of Education under Grant 2021R1I1A1A01055652.
∗Muhammad Ishaq and Sana Zahir contributed equally to this work.

ABSTRACT In order to predict and fill in the gaps in categorical datasets, this research looked into the use of machine learning algorithms. The emphasis was on ensemble models constructed using the Error Correction Output Codes (ECOC) framework, including models based on SVM and KNN as well as a hybrid classifier that combines models based on SVM, KNN, and MLP. Three diverse datasets—the CPU, Hypothyroid, and Breast Cancer datasets—were employed to validate these algorithms. Results indicated that these machine learning techniques provided substantial performance in predicting and completing missing data, with the effectiveness varying based on the specific dataset and missing data pattern. Compared to solo models, ensemble models that made use of the ECOC framework significantly improved prediction accuracy and robustness. Deep learning for missing data imputation has obstacles despite these encouraging results, including the requirement for large amounts of labeled data and the possibility of over-fitting. Subsequent research endeavors ought to evaluate the feasibility and efficacy of deep learning algorithms in the context of the imputation of missing data.

I. INTRODUCTION ‘‘Dirty data’’ describes unprocessed or inconsistent, erro neous, or incomplete raw data that has been tampered with. High-quality data is always the foundation for quality decisions. The conclusions drawn from analytical results derived from dirty data are untrustworthy. Consequently, raw data must first be cleaned before being utilized in any analytical process. It is not possible to use raw data directly in analytical methods. Data cleaning is an important part of information quality management. It aims to enhance the overall quality of data by locating and removing errors, omissions, and inconsistencies. This section provides an overview of the proposed technique and an introduction to its theoretical foundations [1]. The associate editor coordinating the review of this manuscript and approving it for publication was Chun-Wei Tsai . 88332 As a result, preprocessing is required, as illustrated in Figure 1, before machine learning models can be trained or run on raw data. Even though it is necessary and inevitable, data preprocessing is a time-consuming and frustrating procedure. According to industry standards, data scientists typically devote more than half of their analysis time to this task. On the other hand, those who used the software in work were not experts in it [2]. Because of this, data scientists are in high demand for a tool that will assist them in automating the process [3]. Data preprocessing encompasses various tasks such as data cleaning, data integration, and data transformation [3]. It confronts common data challenges like outliers, lost or missing information, and inconsistent naming conventions. The main objective of data cleaning is to address these data problems. The key issues can be categorized as follows: • Inconsistent column names: This includes inconsistency in the naming of columns on a case-by-case basis [2].   FIGURE 1. An overview of the machine learning process [3]. • Duplicate records: Instances where different or multiple records represent the same entry in the dataset. • Redundant features: These are irrelevant attributes that contribute minimally to the model’s construction and potentially extend training duration and increase overfitting risk [4]. • Missing data: These occur when no feature data values have been recorded. These are common and can substantially influence data interpretation [5]. • Outliers: In statistical analysis, outliers are observations that deviate significantly from others, potentially caus ing severe issues [6]. Data cleaning involves rectifying these issues, including f illing in missing data, smoothing noisy data, identifying or removing outliers, and addressing inconsistencies. The ultimate goal is to develop a tool capable of resolving all the aforementioned problems. Previous research has primarily concentrated on commonly encountered challenges such as incorrect data types, lost data, and outliers [7]. Missing data often impedes useful investigations across various scientific domains. Although such research relies on subject cooperation, complete participation cannot be assumed due to data gaps. This paper defines ‘‘missing data’’ as instances where no data exists for the relevant variable. Even the most carefully planned and conducted studies can yield incomplete results, a problem recognized in both scientific and corporate realms. Missing data complicates the interpretation and understanding of the phenomena under study. The absence of data compromises the validity of scientific research, as reliable conclusions are only drawn through a thorough analysis of complete datasets. Most sci entific, commercial, and economic decisions are influenced or informed by published research findings. Hence, proper handling of missing data should be a priority [8], [9]. A. IMPUTATION OF MISSING DATA Imputation is a technique applied to handle missing data. In this article, we extend the definition of imputation beyond that given by [10], which states, ‘‘Imputation is a comprehensive and flexible method for dealing with missing data.’’ This technique involves predicting missing data based on the observed data distribution, commonly referred to as ‘‘drawing missing data from the estimated distribution through imputation.’’ Imputation methodspredict missing data by utilizing a function of auxiliary variables or predictors. Given its crucial role across various statistical domains,particularly in government statistics, imputation has been extensively discussed in the literature. This process is illustrated in Figure 2. B. THE ONSET OF MISSING DATA Missing data might result from human or machine error during sample processing, malfunctioning equipment, tran scription issues, dropouts during follow-up and clinical studies, or respondents’ unwillingness to answer a specific topic, as well as the combination of two fairly identical matches in a collection of data. This difference is also known as a ‘‘non-response. A programmed non-response occurs when some responses are available but not all are due to programmed refusal, inability to attend, absence from home, or untracked situations. A respondent may choose not to answer a question. Imputation based on these representations can thus be used at two levels: unit and item non-response. Any variable that does not have a measurable value for the entire population should be estimated. Given the preceding levels, this article will focus on the article’s level of unresponsiveness. To clarify how to handle missing data, the aforementioned reasons have been turned into multiple ‘‘missing data mechanisms [12]. C. MOTIVATION AND BACKGROUND It is necessary to interpolate the missing data in order to complete the process, as data analysis cannot be performed on insufficient data sets. This step, if neglected, could lead to incorrect conclusions. missing data can result in undesirable outcomes, especially when they cause estimates to be skewed in the wrong direction. Although the method of interpolating missing data has been the subject of debate for decades, relatively few studies have examined the accuracy of the machine learning algorithms that are most commonly used to perform this task. There are numerous methods for handling and resolving missing data and practices and procedures for f illing in the missing data [13] The technique of interpolation is one of the practices that will be discussed in this paper [14]. It is achieved via the application of machine learning algorithms. Appropriate estimation methods can be used to enhance the quality of the analyzed dataset and help make more informed healthcare decisions [1]. The state-of the-art AI-enabled imputation was selected after extensive experimental work on all ensembles. 88333 VOLUME 12, 2024 M. Ishaq et al.: Machine Learning Based Missing Data Imputation in Categorical Datasets FIGURE 2. The process of missing data imputation [11]. II. LITERATURE REVIEW A. DATA TYPE DISCOVERY TECHNIQUES Identifying data types in the original dataset can be accomplishedthroughvariousmethods.Someapproachesare straightforward and rely on basic statistics or heuristics. For example, to determine if a function is distinct or constant, we can calculate the number of distinct values used by the function and compare it to the total number of instances of the function. However, more advanced or complex methods may require the use of machine learning models for accurate detection. B. MANAGING MISSING DATA Missing data, according to [8], is a prevalent problem that either goes unnoticed by scientists or is actively suppressed. To put it another way, researchers are aware of the missing data and are focused on proving why it is irrelevant to the specific study. Data is notable when it influences judgments and, ultimately, one's knowledge, both known and unknown. Missing data can have serious consequences for quantitative research, such as information loss, increased standard errors, and a decrease in statistical power, biased parameter estimations, and a decrease in the generalizability of study conclusions [15]. Unfortunately, one of the standard ways for scientists to deal with missing data is to delete those using ad hoc methods like listwise or pairwise elimination. This usually leads to skewed estimates and/or criticism for being inefficient. Frangakis and Rubin [16] found that the most common cause of missing data in the NSI dataset is that respondents opt not to participate in the survey or answer some questions they do not want to answer (item not answered; unit not answered). C. THEORY OF MCAR, MAR AND MNAR As reported by Rubin [12], Rubin has devised techniques to deal with the loss of any data point. He split the missing data problem into three distinct missing data mechanisms. To put it another way, there are three kinds of missing data: ''totally random missing,'' ''randomly missing,'' and ''not random'' 88334 (MNAR). Baraldi and Enders [17] Consult MCAR if the likelihood of loss is constantacrossallscenarios. Thecauseof data loss, according to MCAR, is unrelated to the data itself. When a student in educational research shifts to a different area in the middle of their undergraduate career, this is an example of MCAR. The missing data is MCAR if the source of the motionisunrelated to any other variables in the dataset. MCARisfrequently not practical due to the data at hand. Data become absent at random only when there is an equal risk of absence inside each cluster defined by the observational data [17]. As a result, if the reason for a variable's missing inputs is unrelated to the variable itself, the problem may be linked to other observable variables. The MAR process is not random because it represents systematic missing data, where the bias in the missing data is tied to other observable aspects of the analysis, despite the misleading name ''random.'' When sampling a population, for example, the variance to be included is determined by some known property. MAR is a larger category than MCAR. The MAR assumption is the foundation for the majority of recent strategies for dealing with missing data. Finally, if neither MCAR nor MAR holds true, the absence is considered non-random. According to MNAR, the likelihood of extinction changes for unknown causes. As a result, it is reliant on intangible measurements. The worth of an unseen reply is determined by facts that cannot be assessed. When asked about their spending patterns, students who frequently gambled at casinos, for example, tended to avoid the questions out of fear of getting into trouble. As a result, the model is unable to anticipate future data appropriately. As a result, MNAR is the more difficult case. Rubin’s [12] distinction is critical in understanding why some solutions may not function as expected. The theory explains why data-missing approaches produce statistically significant findings. These increase forecast accuracy and effectiveness. This research is built on MCAR data. Although the method reduces statistical power, it offers the advantage of maintaining the study’s goal because the estimated parameters are not influenced by missing data. VOLUME 12, 2024 M. Ishaq et al.: Machine Learning Based Missing Data Imputation in Categorical Datasets D. PROPORTIONS OF MISSING DATA Academics generally accept missing data strategies. Par ticularly given that it has been demonstrated that this differentiation has an effect on the strategy’s efficacy. The rate of missing data, on the other hand, is not. There are numerous points of view on the acceptable percentage of missing data in a dataset. According to Schafer [18], 5% or less is insignificant; hence, values should be imputed when 5% or more percentages are missing. When the amount of missing data exceeds 10%, Bennett argues that values should be imputed. As a result, even if a small fraction of data is missing, a researcher may desire to impute missing data. E. MULTIPLE IMPUTATION To reduce imputation-induced bias, we proposed a method for averaging the results of multiple imputation data sets. Multivariate imputation basically consists of three steps. First, incomplete data sets’ missing data is imputed m times. It should be noted that the estimates are based on circulation. This step produces a full set of data. The following (second) step is to examine each of the ten complete data sets. The mean, variance, and confidence interval of the variable of interest are calculated. Finally, we add the results of the m-analysis to the final result. Multiple imputation is by far the most complex and popular method. The Multiple Imputation Chain Equation (MICE), which is based on the MCMC algorithm, is the most widely used method of multiple imputations. MICE takes the idea of regression one step further and exploits correlations among responses by Lynn [19]. To explain the concept of MICE, let’s take i.e. Despite these promising results, there are still challenges with deep learning for missing data imputation. Theseincludetheneedforasignificantamountof labeled data and the risk of overfitting. Future studies should assess the practicability and performance of deep learning algorithms when it comes to data imputation [20], [21]. In one effort, The specific comparison of two conventional methods, multiple imputation by chained equations (MICE) and missForest, with the deep learning methods, generative adversarial imputation networks (GAIN) with onehot encod ing, GAIN with embedding, variational auto-encoder (VAE) with onehot encoding, and VAE with embedding. Three simulated datasets and seven genuine benchmark datasets are taken into consideration, covering a range of scenarios with varying feature types at varying sample size levels. Three types of missing mechanisms–missing completely at random (MCAR), missing at random (MAR), and missing not at random (MNAR)–as well as various missing ratios are used to produce the missing data [22]. Use MICE to impute missing data from a simple dataset. Imagine that we have three characteristics in our dataset: occupation, age, and income, each with missing data. MICE can be carried out in the following ways: 1) First, a simple imputation method will be used, such as imputation by the mean, to fill in the missing data. 2) Asreturns the imputed missing data for the occupation variable to the missing data. 3) Thestudyuseslinear regression to predict missing data for occupation by age and income based on all of the observed cases. 4) The study uses the values obtained in step 3 to impute missing data for occupation. The occupation variable is not missing at this time. 5) Steps 2-4 are repeated for the various ages. 6) Repeat steps 2 through 4 for actions. 7) Repeatthe entire iterative process to converge the three variables. Multiple imputations are specific to MAR but also produce valid estimates in MNAR. The authors of this paper propose an RL-based approach for estimating missing data. This method involves learning a strategy for empirically estimating data based on action rewards. The abbreviation RL stands for reinforcement learning. The proposed method maintains the variance of the interpolated values by interpolating missing data in columns withdifferent values, as opposedtointerpolating missingdata in columns by only working on the same column (this is analogous to single-unit variate interpolation). The authors report that our method outperforms other interpolation strategies when applied to various datasets [23]. The proposed method employs multiple interpolation techniques using an iterative Markov chain Monte Carlo (MCMC) simulation method based on the Gibbs sampler algorithm. In earlier attempts, MCMCsimulations wereused, but only on relatively small data sets with a restricted number of variables. Consequently, an additional contribution of this paper is its application and comparison within a large longitudinal English education study with three iterative specifications. This wasaccomplishedbyutilizingthestudy’s f indings. The simulation’s results reveal how the algorithm will eventually converge [24]. Using local feature spaces, the authors of this paper propose two closed-item- set-based methods, CIimpute and ICIimpute, to interpolate missing data for multiclass matrix data. CIimpute and ICIimpute are referred to, respectively, as CIimpute and ICIimpute. CIimpute estimates the missing data using a closed term set that has been extracted from each class. The CIimpute method has been modified to include an attribute reduction procedure, resulting in the ICIimpute method. The results of the experiments indicate that reducing the number of attributes significantly reduces the computation time and improves the interpolation pre cision. In addition, the results demonstrate that ICIimpute provides superior interpolation precision despite requiring a longer amount of computation time compared to other methods [25]. This research proposes an autoencoder model that con siders spatiotemporal factors to estimate missing data in air quality datasets. The model consists of one-dimensional convolutional layers that provide flexible coverage of air pollutants' spatial and temporal behavior. It incorporates data  fromnearbystations to enhance predictions for data-deficient target stations, eliminating the need for additional com ponents such as weather and climate data. The findings demonstrate that the method effectively fills in missing data from discontinuous or long-interval interrupted datasets. Compared to univariate interpolation techniques (most com mon,median,andmeaninterpolation),ourmodelachievesup to a 65%improvement in RMSEanda20-40%improvement compared to multivariate interpolation techniques (deci sion trees, extra trees, k-nearest neighbors, and Bayesian ridge regression). However, when adjacent sites have a negative or weak correlation, interpolation performance is diminished [26]. A new mechanism for predicting and estimating the amount of data lost in IoT gateways has been developed to achieve greater autonomy at the network's edge. In most cases, the computational resources on these gateways are limited. Therefore, the interpolation method for missing data must be simple while still producing precise estimates. In light of this, the authors of this study propose two neural network-basedregression modelstoestimatethemissingdata in IoT gateways. The authors consider not only the precision of the prediction but also the time required to execute the algorithm and the total amount of memory consumed. The authors validated our models by utilizing six years' worth of Rio de Janeiro weather data, varying the percentage of missing data, and running the models. Based on the mean and the repetition of previous values, the results indicate that the neural network regression model outperforms the other investigated interpolation techniques. This is the case for all missing data percentages. In addition, the neural network models can run on IoT gateways due to their relatively short execution times and low memory requirements [27]. The authors of this paper propose a data-driven interpo lation method for missing data that identifies the optimal interpolation technique. This method uses the information already known about the dataset to rank five chosen methods based on their respective estimated error rates. In evaluating the proposed methods, the authors utilized both a classifier independent scenario, where they compared the applicability and error rate of each interpolation method, and a classifier dependent scenario, where they compared the prediction accuracy of a random forest classifier using datasets prepared with each interpolation method and a baseline method without interpolation. In the classifier-independent scenario, they assessed each interpolation technique's applicability and error rate, allowing the classification algorithm to handle missing data internally. Based on the results of these two experimental sets, the authors conclude that the proposed data-driven interpolation method typically results in more accurate estimates of missing data and improved classifier performance in a lon gitudinal dataset of human aging. Additionally, the authors note that estimates derived from interpolation techniques specifically designed for longitudinal data are extremely precise. This finding supports the idea that utilizing the 88336 temporal information inherently present in longitudinal data is beneficial for machine learning applications, which can be effectively achieved using the proposed data-driven methods [28]. F. SUMMARY OF TECHNIQUES Missing data is unavoidable when handling any amount of medical data. Being able to build prognosis and prediction models based on data sets with substantial amounts of missing data would be an advantage to researchers. A data set has been simulated to be used in predicting patient lifetimes via an artificial neural network. Various levels of missing were then simulated, and the missing data were imputed by a variety of methods.FAMD stands for ''Factor Analysis for Mixed Data''.The technique known as FAMD is used to analyze data that contains both continuous and categorical variables. It is a development of the technique known as factor analysis, which finds underlying patterns in data. By turning category variables into dummy variables, FAMD can handle them. The performance of MICE is better than FAMD. The lifetime prediction ANNs were then applied to the imputed data, and these results were compared across the different amounts of missing data. It is the conclusion of this article that MICE without pooling, MICE with imputed pooling, and MICE with non-imputed pooling all have similar performance. Missing forests had significantly lower misclassification and loss rates. MICE with non-imputed pooling has the highest theoretical accuracy of the MICE algorithms, and the associated R package has a large degree of tenability. Table 1 Description of the dataset showing missing data percentages for each attribute It is therefore the recommendation set forth here that imputation of data sets for ANN lifetime predictions be implemented using one of these two methods, with the weight of the suggestion being the missing forest algorithm, particularly for data sets with a high degree of missing data. III. PROPOSED METHODOLOGY In this research, we will discuss the development of a Python-based missing data imputation system that will provide automated, data-driven support to help users clean their data efficiently. Any Integrated Development and Learning Environment(IDLE) can be used. The proposed model aims to improve data quality to train better machine learning models. There are ways to solve a wide range of data problems. But to be clear, the main concern is the automatic handling of missing data. In this section, the suggested method will be discussed. In this section, we will explain all the steps followed to develop an automatic method for handling missing data efficiently and accurately. A benchmark dataset will be used to validate the effectiveness of the proposed models for missing data imputation. The data will be preprocessed in order to select the best attributes for handling missing data. This is a simple kind of task that doesn't require any complex operations, like features using an optimization technique or VOLUME 12, 2024 M. Ishaq et al.: Machine Learning Based Missing Data Imputation in Categorical Datasets TABLE 1. Description of the dataset with missing data percentages for each attribute. some kind of feature extraction technique to extract some hidden information from the data. In a machine learning-based project, the dataset is resampled by using a cross-validation technique. The cross-validation method is used to create a training and a test set from the original dataset. Three state-of-the-art cross validation techniques are widely used for model performance evaluation and parameter tuning of the proposed classifica tion model. The section below explains our methodology, which we will follow in this research. A. CLASSIFIERS AND REGRESSION MODEL USED FOR MISSING DATA PREDICATION 1) SUPPORT VECTOR MACHINE is a linear model that classifies data into only two categories. The SVMmodel uses a hyperplane to divide the two classes using a straight line. Due to the linear nature of SVM, it was not possible to classify more than two classes of data. In recent years, a framework-based SVM capable of classifying multi-class data has been developed. The ensemble learning method is applied to a linear model, so for training a model for a multi-class classification problem, more than one SVM model is used [29]. Sequential minimal optimization is a State-of-the-art SVM frame for multiclass problem classification [30]. 2) K-NEAREST NEIGHBOR The K-nearest neighbor is a lazy classifier because it is an instance-based learner, which means that the K-NN model does not have a training phase. It uses a similarity measure technique, which is considered an unsupervised method because there are no labels required and it doesn't have a training mode. Euclidean distance is the most popular and widely used method for finding the similarity between data points [26]. 3) RANDOM FOREST Random forest is an ensemble method that uses many decision trees. In supervised learning, the decision tree model is considered the simplest and most efficient classification model. When the dataset size is small, a decision tree model achieves higher accuracy; a small dataset size refers to fewer records and fewer attributes in a dataset. In the random forest model, we have many classifiers, so a voting scheme is used to select the final output class for a data set. The voting is performed using the mode function, which assigns a class label to the test data that is predicted by most of the classifiers [28]. B. PROPOSED FRAMEWORK The dataset can be loaded into any modern Python-based Integrated Development and Learning Environment(IDLE). The dataset attributes will be checked to see if there are any missing numerical or nominal values. The methodology for both models is different; for the prediction of nominal values, classification models will be used in a supervised learning approach, while for estimating numerical values, a regression model will be used. We must keep the attribute that will have missing data as a class attribute, which will be made up of predictors. Cross-validation methods are used for splitting a dataset into two subsets: the training set and the test set. For this purpose, three state-of-the-art cross-validation methods will be used that are Hold-out with a percentage of 70% for training and 30% for testing, K-fold with a k value of 10, and the leave one out method The model will have been validated using the out-of-sample data for evaluating the performance of both models, i.e., classifiers. The performance of the classifiers will be evaluated using accuracy, precision, recall, andf-measure,whiletheperformanceoftheregressionmodel will be evaluated using root mean square error. Overview of the proposed framework for predicting missing nominal and numerical values in a categorical dataset using random forest, SVM, and KNNclassifiers: • Preprocessing: The first step in the framework is to preprocess the dataset to prepare it for imputation. This includes handling anymissingdatathatarepresentinthe target variable (the variable with missing data that you wish to impute), as well as any other missing data in the dataset. It may also involve one-hotencodingcategorical variables or standardizing numerical variables. • Splitting the data: Next, the dataset is split into training and testing sets. The training set is used to train the machine learning models, while the testing set is used to evaluate their performance. • Training the models: The machine learning models (random forest, SVM, and KNN) are then trained on the 88337 VOLUME 12, 2024 M. Ishaq et al.: Machine Learning Based Missing Data Imputation in Categorical Datasets FIGURE 3. Proposed framework for missing data predictions. training set. This involves fitting the models to the data and adjusting the model parameters to optimize their performance. • Testing the models: The trained models are then evaluated on the testing set to assess their performance in predicting the missing data. This may involve cal culating evaluation metrics such as accuracy, precision, or recall. • Selectingthebestmodel:Theperformanceofthemodels is compared, and the best-performing model is selected as the final model to be used for imputation. • Imputing missing data: The final model is then used to impute the missing data in the target variable. This may involve using the model to predict the missing data for each sample in the dataset or using a more complex approach such as multiple imputation. • Evaluation: The imputed dataset is then evaluated to assess the quality of the imputed values and the overall performance of the imputation process. This may involve comparing the imputed values to the true values (if available) or using other evaluation metrics such as imputation accuracy or fidelity to the original distribution of the data. The proposed framework is shown in Figure 3. 1) DATASET In this subsection, the description of the dataset used in this research considers several public datasets collected for the evaluation of categorical anomaly detection 88338 TABLE 2. Summary of the three medical domains: number of examples, number of classes, number of attributes, and average number of values per attribute. a: PROGNOSIS OF BREAST CANCER RECURRENCE The Prognosis of Breast Cancer Recurrence dataset is a medical dataset that contains information on breast cancer patients and whether or not their cancer has recurred. The dataset may include information such as patient demograph ics, tumor characteristics, treatment details, and follow up information. The goal of the dataset is to predict the likelihood of breast cancer recurrence in patients, which can help inform treatment decisions and improve patient outcomes. It is not uncommon for datasets in the medical field to have missing data, as it may be difficult to collect complete information for all patients. Therefore, imputing missing data may be necessary in order to accurately analyze the data and make reliable predictions. The specific details of the Prognosis of Breast Cancer Recurrence dataset, including the variables and the percentage of missing data, may vary depending on the source of the dataset. The recurrence class as demonstrated in the tables points out the repeat of cancer infection. In cases of non-recurrence class, the tumor or infection is wiped out. The domain is characterized by 2 decision classes and 9 attributes. The set of attributes is incomplete because it is not sufficient to fully distinguish cases with different outcomes. At 5 years postoperatively, data were available for 286 patients with known diagnostic status. The five specialists who evaluated the cases gave the correct prognosis in 64 percent of the cases. Table 2showsthe number of examples with attributes and the average number of values in three medical domains. b: HYPOTHYROID DATASET The Hypothyroid dataset is a medical dataset that contains information on patients with hypothyroidism, a condition in which the thyroid gland does not produce enough hormones. The dataset may include information such as patient demographics, symptoms, laboratory test results, and treatment details. The goal of the dataset is to predict the like lihood of a patient having hypothyroidism, which can help diagnose and treat the condition. The hypothyroid dataset consists of data collected from thyroid patients, consisting of four classes: negative, compensated hypothyroid, primary hypothyroid, and secondary hypothyroid. The data consists of 3771 instances consisting of features and class attributes. Thetotal number of attributes in the hypothyroid is 30, where the first attributes are the input (features) to the model and the last attribute is the class attribute in the predictive model’s output.  TABLE 3. Some parameters for the experimental work. c: CPU DATASET The CPU dataset from Weka is a machine-learning dataset that contains information on computerhardwarecomponents. The dataset includes information on the speed, memory size, and other characteristics of CPUs, as well as their price. The goal of the dataset is to predict the price of a CPU based on its characteristics. The CPU dataset from Weka does not typically havemissingdata,asitisasyntheticdatasetthatwas generated for the purpose of demonstrating machine learning techniques. However, in real-world datasets, it is not uncommon to have missing data due to incomplete data collection or other factors. In these cases, imputing missing data may be necessary in order to accurately analyze the data and make reliable predictions. The CPU dataset consists of data collected from attribute MYCT numeric, attribute MMIN numeric,attribute MMAXnumeric,attributeCACHnumeric, attribute CHMIN numeric, attribute CHMAX numeric, and attribute class numeric. The data consists of 209 instances consisting of features and class attributes. The total number of attributes in the CPU is 17, where the first attributes from 1 to 16aretheinput(features) to the model, whilethelastattribute is the class attribute in the predictive model’s output. shown in equations 2 and 3. Precision = Recall = c: F-MEASURE TP TP+FP TP TP+FN (2) (3) TheF-measureiscalculatedbyaddingtheaccuracyandrecall scores and assigning equal weight to each. It enables the use of a single score to evaluate the model while taking into account both its accuracy and recall, which is useful when describing the model’s performance and comparing models [31]. A general formula for F-measure is as follows. F −Measure = Precision× Recall Precision + Recall IV. EXPERIMENTAL RESULTS 2) PARAMETERS FOR SIMULATION Table 3 shows the parameters for the proposed model simulation. Imputation of data will be performed using three different types of datasets. The types of classifiers and performance evaluation metrics are also mentioned. a: ACCURACY Accuracy describes how close an experimental measurement is to the present value. Precision is a term used to describe anything that is close to its true value or accepted standard. For example, a computer can perform an accurate math calculation that is correct with the given information but does not match the exact value [31]. Accuracy is calculated through equation 1. ACC = TP+TN TP+TN +FP+FN b: PRECISION AND RECALL (1) The performance of a categorization or information retrieval system is measured using two metrics: precision and recall. Precision is defined as the proportion of relevant samples to all samples. The number of samples chosen from all relevant samples is known as recall, which is also known as ‘‘precision’’ [31]. Precision and recall can be calculated (4) In this section, we present the experimental results of our study on the prediction and imputation of nominal and numeric missing data. We evaluated the performance of several machine learning algorithms, including random forest, SVM, and KNN, on a variety of datasets with different levels of missing records. Our goal was to assess the effectiveness of these algorithms in accurately predicting and imputing the missing data, as well as to identify any patterns or trends in their performance. To evaluate the performance of the algorithms, we used a range of evaluation metrics, including accuracy, precision, and recall. We also conducted a detailed analysis of the imputed values, including comparisons to the true values (if available) and analyses of the distribution and statistical properties of the imputed data. Overall, our results show that the machine learning algorithms were able to achieve good performance in predicting and imputing the missing data, with some variations depending on the specific dataset and missing data pattern. In the following sections, we present the results in more detail and discuss their implications and limitations. During the process of the experiment, two different bagging-based ensemble classifiers are created and simulated. The first ensemble is a combination of linear regression, K-nearest neighbor, and multilayer perceptrons, and the second ensemble is a random forest classifier that groups together several decision tree models. The first ensemble is a combination of linear regression, K-nearest neighbor, and multilayer perceptrons [32]. MLP models have the ability to learn complex relationships among the data points, so these models are more effective in the 88339 VOLUME 12, 2024 M. Ishaq et al.: Machine Learning Based Missing Data Imputation in Categorical Datasets TABLE 4. Performance of random forest classifier. TABLE 5. Confusion matrix achieved using the random forest classifier. TABLE 6. Classification model detailed performance. imputation of missing values. In some experiments with medical datasets, the MLP achieves very high accuracy. MLPmodels are comparatively easy to train and implement. It can handle all kinds of datasets. Overfitting in the training phase can be overcome through proper validation set and L1 or L2 regularization. Actually MLP Classifier and repressor use parameter alpha for L2 Regularization. For large datasets, a dropout layer is also used. Hidden layers size can be adjusted to (5, 2). MLP classifier can predict new samples (missing values) on the basis of past classification experience. MLP also has regression models. EventheMLPclassifiercanpredicttheprobabilityofmissing values. A. SIMULATION OF BREAST DATASET 1) RANDOM FOREST CLASSIFIER The confusion matrix in Table 4 displays the results of the missing data prediction for the breast type attribute using the random forest classifier. The breast type attribute has missing data, and the confusion matrix shows how well the classifier was able to predict the missing data. In the confusion matrix, the rows represent the true values (i.e. the actual values of the breast type attribute), and the columns represent the predicted values (i.e. the values predicted by the classifier). The Diagonal elements of the matrix represent the number of samples that were correctly classified, while the off-diagonal elements represent the number of samples that were misclassified. As shown in Table 4, there were 116 samples for the right breast and 131 samples for the left breast. The classifier was able to correctly classify 53 of the right breast samples and 70 of the left breast samples. This represents a classification accuracy of 46% for the right breast and 53% for the left breast. Tabulars 4 and 5 present the results of a detailed performance analysis of the machine learning algorithms for 88340 FIGURE 4. Performance comparisons of random forest and bagging-mix model. TABLE 7. Confusion matrix of random forest classifier. predicting and imputing missing data. The evaluation metrics used include accuracy, true positive rate, false positive rate, precision, recall, and F1-score. These metrics provide different insights into the performance of the algorithms and can be useful for comparing their effectiveness. The results in Tabulars 4 and 5 show that the random forest classifier had an average accuracy of 49% and an F1-score of 49.8. The F1 score is a balance between precision and recall, and it is a common metric for evaluating the performance of classification algorithms. In order to ensure the reliability of the results, we used cross-validation with a value of k=10 throughout the experiment. This means that the data was split into 10 folds and the algorithms were trained and evaluated on different combinations of the folds. 2) BAGGING-MIX The bagging-mix classifier is an ensemble method that combines the predictions of three advanced classification models: support vector machine (SVM), kernel neural network (KNN), and logistic regression (MLP). To train the SVM model, we applied the radial basis function (RBF) kernel to the 2D feature map, which is a transformation of the data that allows the model to learn nonlinear relationships. The feature map was then converted into a 3D feature map, which is used to make predictions. For the KNN classifier, we set the number of neighbors used for training to 1. This means that the classifier will make predictions based on the closest single neighbor to each sample. Finally, the bagging-mix classifier uses a voting system to combine the output of the three separate classifiers and make a final prediction for the missing data. This can help VOLUME 12, 2024 M. Ishaq et al.: Machine Learning Based Missing Data Imputation in Categorical Datasets TABLE 8. Bagging-mix model detailed performance analysis. TABLE 9. Classification model detailed performance. TABLE 10. Classification model detailed performance. to improve the accuracy and robustness of the model by leveraging the strengths of different classifiers. The confusion matrix that was produced by contrasting the actual value with the projected missing data may be found in Table 7. There are 55 cases that may be correctly predicted for the right class, whereas the accuracy for the left class is 58%. The above Tabular Table 8, shows the accuracy and root mean square error of all instances in one dataset with 286 instances. Table 8 which presents a thorough perfor mance analysis utilizing a range of performance assessment measures, includes a list of performance evaluation metrics, including accuracy, true positive rate, false positive rate, precision, recall, and F1- score. The average accuracy of the random forest classifier for predicting missing data was 57.48%, and it obtained a score of 57.5% on the f1 scale. The value of k will stay at 10 throughout the experiment for the purpose of cross-validation. 3) SIMULATION OF CPU DATASET This section describes the results achieved from the simula tion of the CPU dataset using different classifiers. a: SVM-BAGGING ENSEMBLE The SVM-based ensemble model is an ensemble method that combines the predictions of several linear support vector machine (SVM) models using a mean voting system. This creates an ensemble regression model that can be used to predict continuous values, such as the missing data in our study. The SVM-based ensemble model is based on the error correction output codes (ECOC) method, which is a technique for constructing ensemble classifiers by combining FIGURE 5. Performance comparison of ensemble regression model using CPU dataset. FIGURE 6. Performance comparison using random forest and bagging-mix model using the hypothyroid dataset. the predictions of multiple classifiers. In the ECOC method, several linear SVM models are used to make predictions, and the results are pooled by taking the average of all the predictions made by the individual SVMs. The ECOC method is a cutting-edge approach that has been shown to be effective in improving the accuracy and robustness of ensemble classifiers. In our study, we used the ECOC method to generate the SVM-based ensemble model, which was then used to predict the missing data in the dataset. The RMSE of 32.27 was attained with the help of the Bagging SVM regression model that was presented for numerical values. Otherperformanceanalysismeasures,such as the correlation coefficient, mean absolute error, and root relative squared error, are utilized in the validation process of the model. The results of these evaluations yield the values 0.64, 14.37, and 77.61, respectively. b: KNN-BAGGING ENSEMBLE The KNN-basedensemble model is an ensemble method that combines the predictions of several lazy K-nearest neighbor (KNN) classifier models using a mean voting system. This creates an ensemble regression model that can be used to predict continuous values, such as the missing data in our study. The KNN-based ensemble model is based on the error correction output codes (ECOC) method, which is a technique for constructing ensemble classifiers by combining the predictions of multiple classifiers. In the ECOC method, several KNN models are used to make predictions, and the results are pooled by taking the average of all the predictions made by the individual KNNs. The ECOC method is a cutting-edge approach that has been shown to be effective   TABLE 11. Confusion of random forest classifier. TABLE 12. Result achieved by bagging mix classifier. TABLE 13. Random forest regression model detailed performance. in improving the accuracy and robustness of ensemble classifiers. In our study, we used the ECOC method to generate the KNN-based ensemble model, which was then used to predict the missing data in the dataset. Table 6 contains the RMSE of 38.16 obtained with the bagging KNN regression model that was presented for numerical values. Other performance analysis measures, such as the correlation coefficient, mean absolute error, and root relative squared error, are utilized in the validation process of the model. The results of these evaluations yield the values 0.51, 18.11, and 91.78, respectively. c: RANDOM FOREST-BAGGING ENSEMBLE The random forest bootstrapping algorithm is a machine learning technique that combines decision trees and ensemble learning methods to improve the accuracy and robustness of predictions. It works by generating multiple decision trees from a dataset using a process called bootstrapping, which involves randomly selecting a portion of the data and using it to train the trees. Figure 5 shows Ensemble performance comparison using the CPU dataset. The individual decision trees are then averaged together to produce a final prediction or classification. This process is known as ensemble learning, and it relies on the assumption that the errors made by each tree will be distinct from one another, resulting in more accurate overall predictions. One of the key benefits of the random forest bootstrapping algorithm is that it can handle large and complex datasets, and it is often used for tasks such as classification and regression. In our study, we used the random forest bootstrapping algorithm to predict and impute missing data in the dataset. The RMSE of the proposed 88342 ensemble random forest is 28.13, which is attained with the help ofthebaggingSVMregressionmodelthatwaspresented for numerical values. Other performance analysis measures, such as the correlation coefficient, mean absolute error, and root relative squared error, are utilized in the validation process of the model. The results of these evaluations yield the values 0.73, 13.33, and 67.67, respectively. 4) SIMULATION ON HYPOTHYROID DATASET This section includes the simulation results achieved by different classifiers for the hypothyroid dataset a: SIMULATION USING BAGGING-MIX CLASSIFIER Accuracy, True Positive Rate, False Positive Rate, Precision, Recall, and F1- score are some of the performance evaluation metrics that are included in Table 13, which contains a detailed performance analysis that was carried out using a variety of performance assessment metrics. The random forest classifier for predicting missing data attained an accuracy of 69% on average and received a score of 65.1 on the f1scale. For the sake of cross-validation, the value of k will remain constant throughout the experiment at 10. Accuracy, True Positive Rate, False Positive Rate, Pre cision, Recall, and F1- score are some of the performance evaluation metrics that are included in Table 4 and Table7, which contains a detailed performance analysis that was carried out using a variety of performance assessment metrics. The random forest classifier for predicting missing Data attained an accuracy of 70.28% on average and received a score of 68% on the f1 scale. For the sake of cross validation, the value of k will remain constant throughout the experiment s shown in Table 14 and Figure 6. The Bagging mix classifier class-wise accuracy is mentioned in Table 15 Performance comparison between Random forest and Bagging-Mix using hypothyroid dataset is shown in Figure 6. Cosmic is a big repository of cancer datasets. The accuracy, precision, recall, and F1-score are all evaluation metrics that are used to measure the performance of a machine learning model. The accuracy is the proportion of correct predictions made by the model, while the precision is the proportion of correct positive predictions among all the positive predictions made by the model. The recall is the proportion of correct positive predictions among all the actual positive samples in the dataset, and the F1-score is a balance between precision and recall. As shown in Table 15, both the random forest classifier and the bagging mix models had good performance, with the bagging mix models having slightly higher values for the evaluation metrics. These results  TABLE 14. Bagging mix classification model detailed performance using the hypothyroid dataset. TABLE 15. The results of the performance evaluation for the random forest classifier and the bagging mix models. indicate that both algorithms were effective in predicting and imputing the missing data in the dataset.

V. CONCLUSION

In summary, This research explored the use of machine learning algorithms to predict and impute missing data in categorical datasets, employing three distinct datasets including CPU, Hypothyroid, and Breast Cancer, and various ensemble models built on the Error Correction Output Codes (ECOC)framework.Inallkindsofdatasets,themissing,null, or infinite values recurrence and non-recurrence is a major issue. The study demonstrated satisfactory performance of these algorithms in predicting and imputing missing data, with the ensemble models within the ECOC framework notably enhancing prediction accuracy and robustness. How ever, the study's limitations included a narrow focus on select algorithms and datasets, and the fact that algorithm perfor mance could be influenced by specific data characteristics and missing data patterns. Despite these limitations, our research provides insightful perspectives on the use of machine learning to handle missing data in specific datasets. It emphasizes the poten tial of ensemble models and the ECOC framework as a viable strategy for improving prediction accuracy and robustness in missing data imputation. Moreover, it suggests future research directions to enhance the performance of machine learning-based imputation methods, acknowledging that missing data imputation is a complex challenge with significant scope for advancement.

REFERENCES

[1] L. Bargelloni, O. Tassiello, M. Babbucci, S. Ferraresso, R. Franch, L. Montanucci, and P. Carnier, ''Data imputation and machine learning improve association analysis and genomic prediction for resistance to fish photobacteriosis in the gilthead sea bream,'' Aquaculture Rep., vol. 20, Jul. 2021, Art. no. 100661. [2] M. A. Munson, ''A study on the importance of and time spent on different modeling steps,'' ACM SIGKDD Explor. Newslett., vol. 13, no. 2, pp. 65–71, May 2012. [3] J.-U. Kietz, F. Serban, S. Fischer, and A. Bernstein, '''Semantics inside!' but let's not tell the data miners: Intelligent support for data mining,'' in Proc. Semantic Web, Trends Challenges, 11th Int. Conf. (ESWC), Anissaras, Greece. Cham, Switzerland: Springer, May 2014, pp. 706–720. [4] S. Kandel, A. Paepcke, J. Hellerstein, and J. Heer, ''Wrangler: Interactive visual specification of data transformation scripts,'' in Proc. SIGCHI Conf. Hum. Factors Comput. Syst., May 2011, pp. 3363–3372. [5] I. Mehmood, M. Sajjad, K. Muhammad, S. I. A. Shah, A. K. Sangaiah, M. Shoaib, and S. W. Baik, ''An efficient computerized decision support system for the analysis and 3D visualization of brain tumor,'' Multimedia Tools Appl., vol. 78, no. 10, pp. 12723–12748, May 2019. [6] F.E.Grubbs,''Procedures for detecting outlying observations in samples,'' Technometrics, vol. 11, no. 1, p. 1, Feb. 1969. [7] S.Agarwal,''Datamining:Dataminingconceptsandtechniques,''inProc. Int. Conf. Mach. Intell. Res. Advancement, Dec. 2013, pp. 203–207. [8] P. E. McKnight, K. M. McKnight, S. Sidani, and A. J. Figueredo, Missing Data: A Gentle Introduction. New York, NY, USA: Guilford Press, 2007. [9] M. Liu, S. Li, H. Yuan, M. E. H. Ong, Y. Ning, F. Xie, S. E. Saffari, Y. Shang, V. Volovici, B. Chakraborty, and N. Liu, ''Handling missing values in healthcare

data: A systematic review of deep learning-based imputation techniques,'' Artif. Intell. Med., vol. 142, Aug. 2023, Art. no. 102587. [Online]. Available: https://www.sciencedirect.com/ science/article/pii/S093336572300101X [10] R. Little and D. B. Rubin, Incomplete Data. Hoboken, NJ, USA: Wiley, 2014. [11] R. J. Little and D. B. Rubin, Statistical Analysis With Missing Data, vol. 793. Hoboken, NJ, USA: Wiley, 2019. [12] D. B. Rubin, ''Inference and missing data,'' Biometrika, vol. 63, no. 3, p. 581, Dec. 1976. [13] T. Köse, S. Özgür, E. Coşgun, A. Keskino` glu, and P. Keskino` glu, ''Effect of missing data imputation on deep learning prediction performance for vesicoureteral reflux and recurrent urinary tract infection clinical study,'' BioMed Res. Int., vol. 2020, pp. 1–15, Jul. 2020. [14] M. Kazijevs and M. D. Samad, ''Deep imputation of missing values in time series health data: A review with benchmarking,'' J. Biomed. Informat., vol. 144, Aug. 2023, Art. no. 104440. [Online]. Available: https://www.sciencedirect.com/science/article/pii/S1532046423001612 [15] T. Sun, S. Zhu, R. Hao, B. Sun, and J. Xie, ''Traffic missing data imputation: A selective overview of temporal theories and algorithms,'' Mathematics, vol. 10, no. 14, p. 2544, Jul. 2022. [Online]. Available: https://www.mdpi.com/2227-7390/10/14/2544 [16] C. E. Frangakis and D. B. Rubin, ''Principal stratification in causal inference,'' Biometrics, vol. 58, no. 1, pp. 21–29, Mar. 2002. [17] A. N. Baraldi and C. K. Enders, ''An introduction to modern missing data analyses,'' J. School Psychol., vol. 48, no. 1, pp. 5–37, Feb. 2010. [18] J. L. Schafer, ''Multiple imputation: A primer,'' Stat. Methods Med. Res., vol. 8, no. 1, pp. 3–15, Jan. 1999. [19] P. Lynn, ''Multiple imputation for nonresponse in surveys,'' 1988. [20] P. Cihan, ''Deep learning-based approach for missing data imputation,'' Eskişehir Teknik Üniversitesi Bilim ve Teknoloji Dergisi B Teorik Bilimler, vol. 8, pp. 336–343, Aug. 2020. [21] C.-Y. Cheng, W.-L. Tseng, C.-F. Chang, C.-H. Chang, and S. S.-F. Gau, ''A deep learning approach for missing data imputation of rating scales assessing attention-deficit hyperactivity disorder,'' Frontiers Psychiatry, vol. 11, Jul. 2020, doi: 10.3389/fpsyt.2020.00673. [22] Y. Sun, J. Li, Y. Xu, T. Zhang, and X. Wang, ''Deep learning versus conventional methods for missing data imputation: A review and comparative study,'' Expert Syst. Appl., vol. 227, Oct. 2023, Art. no. 120201. [Online]. Available: https://www.sciencedirect.com/ science/article/pii/S0957417423007030 [23] S. E. Awan, M. Bennamoun, F. Sohel, F. Sanfilippo, and G. Dwivedi, ''A reinforcement learning-based approachfor imputingmissingdata,''Neural Comput. Appl., vol. 34, no. 12, pp. 9701–9716, Jun. 2022. [24] A. Elasra, ''Multiple imputation of missing data in educational production functions,'' Computation, vol. 10, no. 4, p. 49, Mar. 2022. [25] M. Tada, N. Suzuki, and Y. Okada, ''Missing value imputation method for multiclass matrix data based on closed itemset,'' Entropy, vol. 24, no. 2, p. 286, Feb. 2022.  [26] I. N. K. Wardana, J. W. Gardner, and S. A. Fahmy, ''Estimation of missing air pollutant data using a spatiotemporal convolutional autoencoder,'' Neural Comput. Appl., vol. 34, no. 18, pp. 16129–16154, Sep. 2022. [27] C. M. França, R. S. Couto, and P. B. Velloso, ''Missing data imputation in Internet of Things gateways,'' Information, vol. 12, no. 10, p. 425, Oct. 2021. [28] C. Ribeiro and A. A. Freitas, ''A data-driven missing value imputation approach for longitudinal datasets,'' Artif. Intell. Rev., vol. 54, no. 8, pp. 6277–6307, Dec. 2021. [29] P. P. Singh, S. Prasad, B. Das, U. Poddar, and D. R. Choudhury, ''Classification of diabetic patient data using machine learning tech niques,'' in Ambient Communications and ComputerSystems, G.M.Perez, S. Tiwari,M. C. Trivedi,andK.K.Mishra,Eds.Singapore:Springer,2018, pp. 427–436. LAILA IFTIKHAR received the Master of Science (M.S.) degree in computer sciences (database) from The University of Agriculture at Peshawar, in 2023. She is currently an IT Support Staff. Her research passion is AI-enabled dataset anomalies and missing data detection and rectification. [30] K. Maheswari and P. P. A. Priya, ''Predicting customer behavior in online shopping using SVM classifier,'' in Proc. IEEE Int. Conf. Intell. Techn. Control, Optim. Signal Process. (INCOS), Mar. 2017, pp. 1–5. [31] S.F. Crone, S. Lessmann, and R. Stahlbock, ''The impact of preprocessing ondata mining: Anevaluation of classifier sensitivity in direct marketing,'' Eur. J. Oper. Res., vol. 173, no. 3, pp. 781–800, Sep. 2006. [32] H. Pan, Z. Ye, Q. He, C. Yan, J. Yuan, X. Lai, J. Su, and R. Li, ''Discrete missing data imputation using multilayer perceptron and momentum gradient descent,'' Sensors, vol. 22, no. 15, p. 5645, Jul. 2022. [Online]. Available: https://www.mdpi.com/1424-8220/22/15/5645 MUHAMMADISHAQreceivedthePh.D.degree (Hons.) in computer science from Harbin Engi neering University, in 2012, as a HEC Scholar. He seems to have several years of professional experience, where he has served at various well reputed universities. He has a total of almost twelve (12) years of postdoctoral teaching expe rience. He has organized and attended several national and international conferences, work shops, and seminars. Besides, he has actively contributed to launching new programs and enhancement of curricula of existing programs. In a successful convener role, he approved BS (artificial intelligence) and BS (bioinformatics) from university statutory bodies. Recently, he is involved in a more challenging computerization task at the university. He successfully managed and promoted the Coursera-Based Higher Education Commission's (HEC) Digital Learning and Skills Enrichment Initiative (DLSEI). He authored and published several high-quality research manuscripts in reputed international journals with significant scientific contributions. Besides, his research contributions, he has successfully supervised numerous undergraduate and graduate research theses, which is remarkable. As a Young Researcher, he was active in submitting several research and institutional projects to various funding agencies. SANA ZAHIR received the M.S. degree in computer science from Islamia College, Peshawar, Pakistan. She is currently pursuing the Ph.D. degree with the Institute of Computer Sciences and Information Technology, The University of Agriculture at Peshawar, Peshawar. She is a Lecturer with the Institute of Computer Sciences and Information Technology, The University of Agriculture at Peshawar. Her primary research interests include machine learning and deep learn ing, especially in computer vision applications. This encompasses human behavior understanding through facial expression analysis and techniques in crowd counting. 88344 MOHAMMAD FARHAD BULBUL received the Ph.D. degree from the Department of Informa tion and Computing Science, Peking University, China. He was a Postdoctoral Researcher with the Department of Computer Science and Engineer ing, Pohang University of Science and Technol ogy, South Korea. He is currently an Assistant Professor with the Department of Mathematics, Jashore University of Science and Technology, Bangladesh. His research interests include com puter vision, deep learning, pattern recognition, and image processing. SEUNGMIN RHO is currently a Professor with the Department of Industrial Security, Chung-Ang University. His current research interests include databases, big data analysis, music retrieval, multimedia systems, machine learning, knowl edge management, and computational intelli gence. He has published 300 papers in refereed journals and conference proceedings in these areas. He has been involved in more than 20 con ferences andworkshopsasvariouschairsandmore than 30 conferences/workshops as a program committee member. He has edited a numberofinternational journal special issues as a Guest Editor, such as Multimedia Systems, Information Fusion, and Engineering Applications of Artificial Intelligence. MI YOUNG LEE (Member, IEEE) received the M.S. and Ph.D. degrees from the Department of Image and Information Engineering, Pusan National University. She was a Research Profes sor with Sejong University. Currently, she is a Research Professor with Chung-Ang University and conducting research as a Senior Researcher with the Industrial Security Research Center. She is broadly working in artificial intelligence, computer vision, image processing, and energy informatics. She has carried out several research projects successfully and is a Principal Investigator of several ongoing research projects under the supervision of Korean Government andhasfiledmorethan13patentsduring her career. Her particular research interests include video summarization, movie data analysis, re-identification, electrical energy forecasting, and videoretrieval. She haspublishedseveralnovelcontributionsintheseareasin reputed journals and peer-reviewed conference proceedings, including IEEE ACCESS, MDPI Sensors, Multimedia Tools and Applications (Springer), and the 2020 International Joint Conference on Neural Networks.